\documentclass[runningheads]{llncs}
\usepackage{graphicx}
\usepackage{multirow}
\usepackage{booktabs,graphicx}
\usepackage[dvipsnames,table,xcdraw]{xcolor}
\usepackage{tikz}
\usepackage{makecell}
\usepackage{comment}
\usepackage{caption}
\usepackage{subcaption}
\usepackage{bbding}
\usepackage{amsmath,amssymb} 

\usepackage[accsupp]{axessibility}  

\begin{document}

\title{Learning Shadow Correspondence for Video Shadow Detection} 


%
\author{Xinpeng Ding\inst{1}\orcidID{0000-0001-7653-1199} \and
Jingwen Yang\inst{1}\orcidID{0000-0003-2420-0130} \and
Xiaowei Hu\inst{2}\orcidID{0000-0002-5708-7018} \and
Xiaomeng Li\inst{1}\orcidID{0000-0003-1105-8083}\thanks{Corresponding Authors}
}
%
%
\institute{The Hong Kong University of Science and Technology
\email{\{xdingaf,jyangbv\}@connect.ust.hk}, \email{eexmli@ust.hk}\\
 \and
Shanghai AI Laboratory \\
\email{huxiaowei@pjlab.org.cn}}
\maketitle

\begin{abstract}

Video shadow detection aims to generate consistent shadow predictions among video frames. 
%
However, the current approaches suffer from inconsistent shadow predictions across frames, especially when the illumination and background textures change in a video.
We make an observation that the inconsistent predictions are caused by the shadow feature inconsistency, \emph{i.e.}, the features of the same shadow regions show dissimilar proprieties among the nearby frames.
%
%
%
%
In this paper, we present a novel \textbf{S}hadow-\textbf{C}onsistent \textbf{Cor}respondence method (\textbf{SC-Cor}) to enhance pixel-wise similarity of the specific shadow regions across frames for video shadow detection. 
%
Our proposed SC-Cor has three main advantages.
Firstly, without requiring the dense pixel-to-pixel correspondence labels, SC-Cor can learn the pixel-wise correspondence across frames in a weakly-supervised manner.
Secondly, SC-Cor considers intra-shadow separability, which is robust to the variant textures and illuminations in videos.
%
Finally, SC-Cor is a plug-and-play module that can be easily integrated into existing shadow detectors with no extra computational cost.
We further design a new evaluation metric to evaluate the temporal stability of the video shadow detection results.
%
%
Experimental results show that SC-Cor outperforms the prior state-of-the-art method, by 6.51\% on IoU and 3.35\% on the newly introduced temporal stability metric.
%
%

%


%

%
\keywords{Shadow detection, video understanding, and correspondence learning.}
\end{abstract}

\section{Introduction}\label{sec:intro}
Shadows in natural images or videos present different colors and brightness.
Known where the shadow is, we can infer light source directions~\cite{lalonde2009estimating,panagopoulos2009robust}, scene geometry~\cite{karsch2011rendering,okabe2009attached,junejo2008estimating}, and camera locations or parameters~\cite{junejo2008estimating}.
Therefore, shadow detection has attracted a lot of attention and achieved remarkable progress.
However, most of the recent methods~\cite{zheng2019distraction,khan2014automatic,vicente2016large,nguyen2017shadow,chen2020multi,hu2020direction,zhu2018bidirectional} detect shadows from single images while shadow detection over dynamic scenes,~\emph{i.e.}, in videos, is less explored.

\begin{figure}[t]
	\centering
	\includegraphics[width=0.85\columnwidth,height=0.33\textheight]{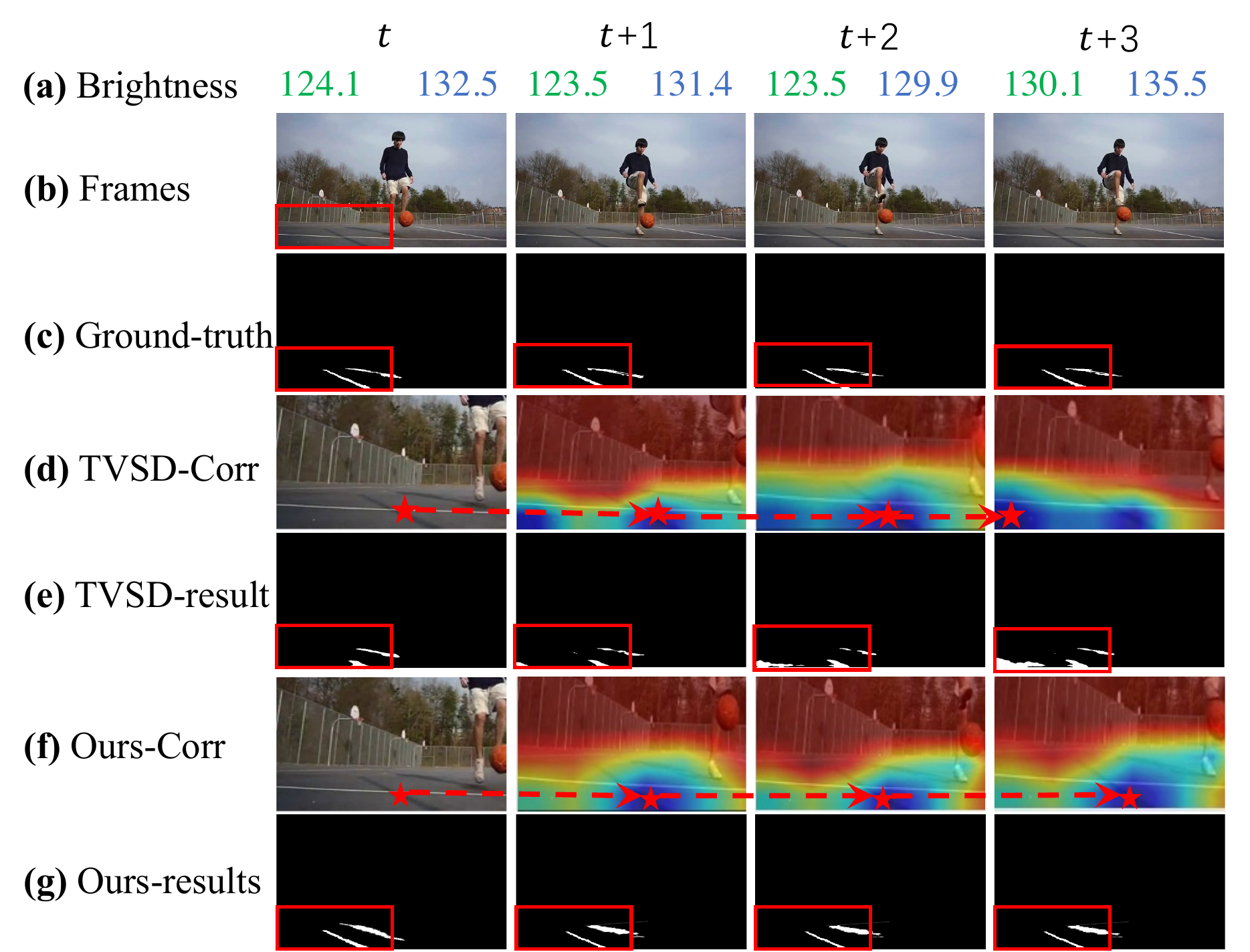}
	\caption{
	Comparison of correspondences and results of TVSD~\cite{chen2021triple} and Ours. We compute
    the brightness of four selected frames. Green and blue values indicate the non-shadow and shadow regions respectively.
    Given a query shadow region in the $t$-th frame,~\emph{i.e.}, the \textcolor{orange}{orange pentagram}, we find the most similar features in nearby frames for the query, and regard the found features as its correspondences.
  ``TVSD-Corr"  and ``Ours-Corr" indicate the correspondences found by TVSD~\cite{chen2021triple} and our method respectively.
  ``TVSD-result" and ``Our-result" refer to the results predicted by TVSD and our method.
  It is clear that the found correspondences in the $(t+3)$-th frame are in non-shadow regions (dark areas).
  This shadow features inconsistency,~\emph{i.e.}, features of the same shadow region may be dissimilar across frames, would result inconsistent prediction (\textcolor{red}{red boxes}).
  Our method can address the shadow feature inconsistency, and generate the contiguous results.
	}
	\label{fig:stable}
\end{figure}

%
To explore the powerful representation capability of deep learning for video shadow detection (VSD), Chen~\emph{et al.}~\cite{chen2021triple} collect a large-scale video shadow detection (ViSha) dataset covering various scenarios. Then, a global contrastive objective is applied on the frame-level, which enhances the similarity between frames in the same video and push away the representations of frames from different videos. 
%
However, video shadow detection is a fine-grained pixel-level detection task, this frame-level semantic constraint may ignore shadow details, i.e., the same shadow
regions across frames show dissimilar, resulting in inconsistent predictions;
see \textcolor{red}{red boxes} in Fig.~\ref{fig:stable}~(d).
%
Video data has the inherent property of \emph{temporal consistency}, where the nearby frames are expected to contain similar shadow regions.
Hence, we aim to explore both frame-level accuracy and temporal-level consistency for video shadow detection.
In this paper, we make a critical observation that this inconsistent prediction is caused by the shadow feature inconsistency,~\emph{i.e.}, the features of the same shadow regions show dissimilar proprieties among the nearby frames.
For example, due to the illumination change (see Fig.~\ref{fig:stable}~(a)), the extracted features in a specific shadow region may show higher similarity with dark non-shadow regions in the nearby frames; see \textcolor{orange}{orange pentagram} in the $(t+3)$-th frame in Fig.~\ref{fig:stable}~(c).
\begin{figure}[t]
    \centering
    \includegraphics[width=1.0\columnwidth,height=0.16\textheight]{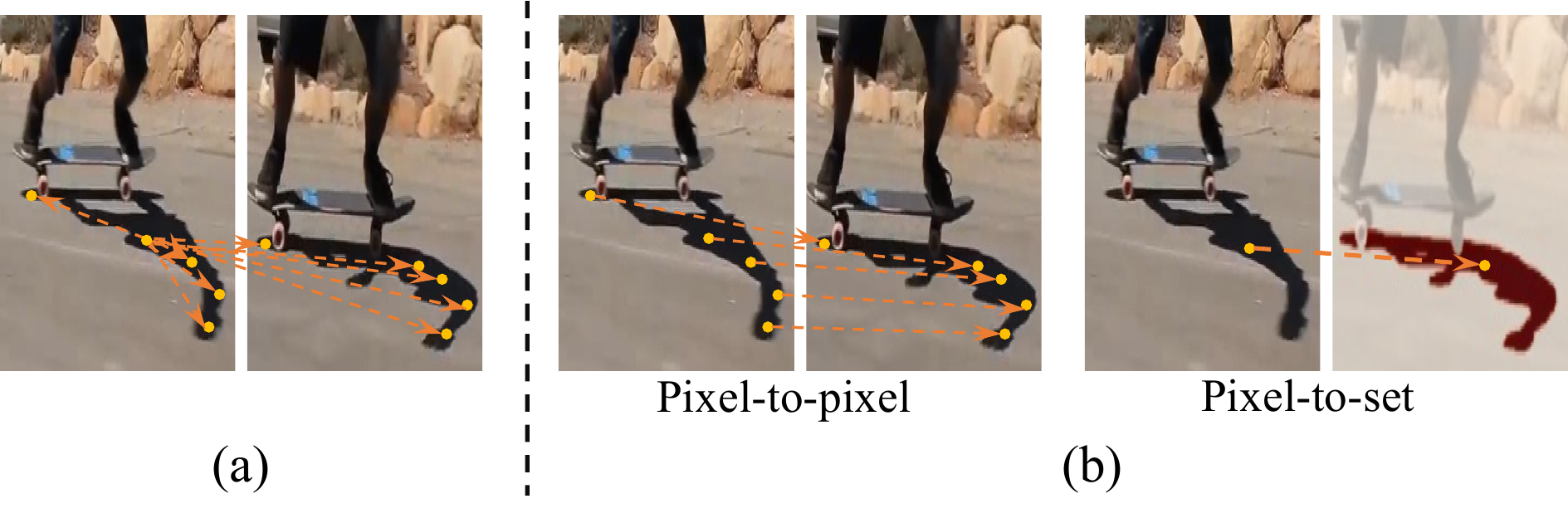}
    \caption{
   (a)Supervised contrastive learning pulls close all pixels in shadow regions, which is too strict to generate complete shadow detection; see Fig.~\ref{fig:corres} and Table~\ref{tab:scl&ours} for details. 
    (b) 
    We aims to leverage correspondence learning to consider intra-shadow  separability.
    Unlike existing correspondence learning~\cite{melekhov2019dgc,jiang2021cotr} that require \emph{pixel-to-pixel} labels among video frames, for each pixel in the shadow, we only know its corresponding pixel is within a shadow region in another frame, denoted as \emph{pixel-to-set} correspondence learning.  
    }
    \label{fig:shadow_corres}
\end{figure}

To address this problem, we aim to enhance temporal pixel-wise similarity for the specific shadow regions across frames, thus improving the detection accuracy and consistency  in shadow videos; see Figs.~\ref{fig:stable}~(e) and (f).
The supervised contrastive learning aims to increase intra-class compactness and inter-class separability, and has been used for image classification~\cite{khosla2020supervised} or semantic segmentation~\cite{zhao2021contrastive,zhang2021looking}.
An intuitive way is to use the supervised contrastive learning to pull close pixels in shadow regions across frames, and push away shadowed pixels and non-shadowed pixels.
However, as objects move and illumination changes, the same shadow region may appear on backgrounds with different textures across frames.
Simply adopting the supervised contrastive learning for video,~\emph{i.e.}, pulling close all pixels in shadows, leads to incomplete shadow regions; see Fig.~\ref{fig:corres} and Table~\ref{tab:scl&ours} for the comparison results.
%

Hence, in this paper, we leverage the correspondence learning to learn a more fine-grained pixel-wise similarity.~\emph{i.e.}, only encouraging a pixel to be similar with its corresponding pixels in nearby frames.
However, unlike the existing correspondence learning~\cite{melekhov2019dgc,jiang2021cotr,ding2021support} that requires \emph{pixel-to-pixel} correspondence labels across video frames, we do not need the pixel-wise correspondence labels. 
%
%
%
%
To this end, we present a novel \textbf{S}hadow-\textbf{C}onsistent \textbf{Cor}respondence method, namely \textbf{SC-Cor}, to learn the dense shadow correspondence in a \emph{pixel-to-set} way, based on a key prior knowledge that a corresponding \emph{(pixel)} of shadow is within a shadow region \emph{(set)} in another frame; see Fig.~\ref{fig:shadow_corres}~(a).   
%
Different from the supervised
contrastive learning~\cite{zhao2021contrastive,khosla2020supervised,zhang2021looking}, our proposed SC-Cor keeps the pixel most similar to the anchor in the shadow region and considers inter-shadow separability, which is robust to the variant textures and illuminations in videos.
%
Note that our SC-Cor is a plug-and-play module and is only used in the training process. Therefore, SC-Cor can be easily applied to any deep-learning-based video/image shadow detection method without additional computational cost in testing.   

Finally, existing metrics only evaluate the performance of VSD in frame-level,~\emph{e.g.}, frame-level Balance Error Rate (BER), and ignores the temporal consistency of shadow predictions.
To this end, we introduce a new evaluation metric, temporal stability (TS), which computes the intersection over union score between the adjacent frames, thus helping to evaluate the temporal consistency of shadow predictions in videos.
%
%
%
%
%
%
Below, we summarize the major contributions of this work:
\begin{itemize}
	%

	\item 
	We present a novel and plug-and-play shadow-consistent correspondence (SC-Cor) method for video shadow detection. Compared with the existing \emph{pixel-to-pixel} learning, our proposed SC-Cor is learned in a \emph{pixel-to-set} way, without requiring pixel-wise correspondence labels.  
	
	
	

	\item  To fairly evaluate the temporal consistency of different shadow detection approaches, we introduce a new evaluation metric, which evaluates the flow-warped IoU between the adjacent video frames.

	\item 
	We evaluate our SC-Cor on the benchmark dataset for video shadow detection and the experimental results show that our method clearly outperforms various state-of-the-art approaches in terms of both frame-level and temporal-level evaluation metrics.
	
\end{itemize}



\section{Related Work}
\noindent{\bf Image shadow detection.}
%
Early traditional methods are based on the hand-crafted shadow features, \emph{e.g.}, intensity, chromaticity, physical properties, geometry, and textures~\cite{sanin2012shadow}.
Recently, deep-learning-based methods become the mainstream algorithms for shadow detection~\cite{khan2014automatic,shen2015structure,vicente2016large,nguyen2017shadow,hu2018direction,hu2020direction,le2018ad,zhu2018bidirectional,zheng2019distraction,ding2019argan}.
%
Khan \emph{et al.} \cite{khan2014automatic} build the first method based on deep neural network, which is a seven-layer CNN that learns from super-pixel level features and object boundaries. 
Hu \emph{et al.} \cite{hu2021revisiting} present a fast shadow detection network by designing a detail enhancement module to refine shadow details.
In the most recent work, Zhu \emph{et al.} \cite{zhu2021mitigate} design a feature decomposition and re-weighting scheme, which leverages intensity-variant and intensity-invariant features via self-supervision to mitigate the susceptibility of the intensity cue.
Except the general shadow detection, Wang \emph{et al.}~\cite{wang2020instance,wang2021single} detecte the shadow regions associated with the objects simultaneously.

\noindent{\bf Video shadow detection.}
Early traditional video shadow detection (VSD) methods adopt the hand-crafted spectral and spatial features \cite{benedek2008bayesian,jacques2005background,nadimi2004physical} to detect the shadow regions.
%
To exploit the capability of deep-learning-based methods on this task, Chen \emph{et al.} \cite{chen2021triple} collect the first large-scale VSD dataset ViSha. 
To detect the shadows in videos, they design a deep-learning-based method that contains a dual gated co-attention module and an auxiliary similarity loss to mine frame-level consistency information between different videos.
Hu \emph{et al.} \cite{hu2021temporal} capture the temporal consistency by an optical-flow-based warping module to align and combine features between video frames.
%
However, due to lack of the temporal pixel-level relation, these methods would suffer from shadow feature inconsistency and generate temporal-inconsistent results.
Unlike existing methods, this paper presents a novel solution to learn pixel-wise consistency by formulating the dense shadow correspondence objective. 
Our method is flexible and can be easily integrated into many existing methods designed for both single-image and video shadow detection methods. 
\noindent{\bf Correspondence learning.}
Finding correspondences between pairs of images is a fundamental task in computer vision~\cite{tyszkiewicz2020disk,sarlin2020superglue,bian2017gms,truong2020glu,melekhov2019dgc,jiang2021cotr}.
However, these methods require pixel-level correspondence labels and can hardly be obtained in videos.
Hence, numerous works aim to learn temporal correspondence in the unsupervised way~\cite{wang2019learning,xu2021rethinking,zhang2020learning}.
These methods perform unsupervised correspondence learning on videos and show obvious improvement on the obvious foreground objects. However, shadows are usually less obvious than the foreground, and may show different appearances and deformation due to illumination and texture changes. Our SC-Cor can address the above problems in a weakly supervised way, which is proved by experiments (see Fig.~\ref{fig:corres}
and Table~\ref{tab:scl&ours}).
In this paper, different from all of these methods, we aim to learn pixel-wise similarity in a \emph{pixel-to-set} way.

\noindent{\bf Contrastive learning.} 
Contrastive learning pulls close an anchor and a positive sample, and pushes the anchor away from many negative samples, which has show great success in self-supervised learning~\cite{chen2021exploring,caron2020unsupervised,he2020momentum,chen2020improved,ding2021support,ding2022exploring,ding2022free}.
Recently, the supervised contrastive learning aims to increase intra-class compactness and inter-class separability to improve image classification~\cite{khosla2020supervised,li2021adaptively} or semantic segmentation~\cite{zhao2021contrastive,zhang2021looking}.
%
%
%
However, as objects move and illumination changes, the same shadow region may appear on backgrounds with different textures across frames. 
The supervised contrastive learning,~\emph{i.e.}, simply pulling close all pixels in shadow regions is too strict, resulting in generating incomplete shadow regions; see Fig.~\ref{fig:corres} and Table~\ref{tab:scl&ours} for details.
Differently, our proposed SC-Cor aims to keep the pixel most similar to the anchor in the shadow regions,
which considers inter-class separability due to the varying shadows in
the videos.
%

\begin{figure}[t]
    \centering
    \includegraphics[width=1.0\columnwidth,height=0.17\textheight]{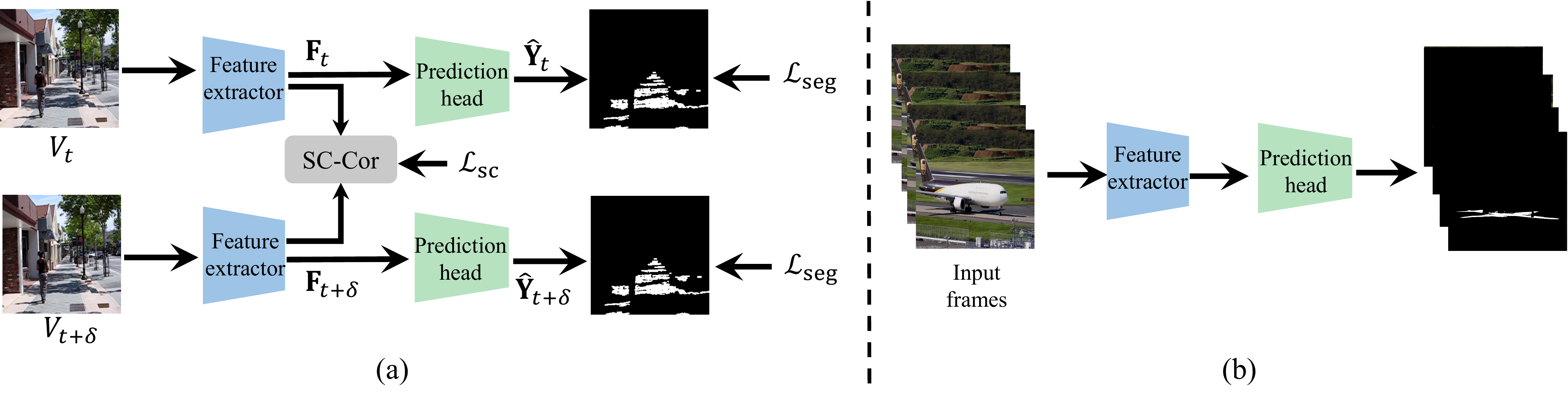}
    \caption{
   (a) Illustration of the training process integrated with our shadow-consistent correspondence (SC-Cor) learning objective.
Given two frames from one video, besides using the the segmentation loss $\mathcal{L}_{seg}$ to supervise their frame-level predictions individually, we also enhance their temporal consistency by SC-Cor, described in Section~\ref{sec:sac}.
   (b) Illustration of the inference phase.
  The proposed SC-Cor is only applied during training. We can improve the temporal consistency as well as the frame-level accuracy without any extra parameters or computation cost during inference.
    }
    \label{fig:architecure}
\end{figure}
\section{Methodology}
%
Fig.~\ref{fig:architecure} (a) shows the training process of the overall SC-Cor framework, which can generate temporal-consistent and accurate shadow detection results.
%
%
Formally, we denote a video sequence and the corresponding ground-truth (GT) masks as $\{ V_t \}_{t=1}^T$ and $\{ \mathbf{Y}_t  \}_{t=1}^T$, respectively, where $T$ is the frame number of this video sequence.
Given two video frames, 
which are denoted as $V_t$ and $V_{t+\delta}$ and $\delta$ is the time interval, we feed them into two branches of the framework; see Fig.~\ref{fig:architecure}~(a).
Each branch contains a feature extractor, which is used to capture the spatial features, \emph{i.e.}, $\mathbf{F}_t$ and $\mathbf{F}_{t+\delta}$ of the input frames.
Note that the weights in these two feature extractors are shared.
Then, we adopt shadow-consistent correspondence (Fig.~\ref{fig:sac}) to extract the temporal information of $\mathbf{F}_t$ and $\mathbf{F}_{t+\delta}$; see details in Sec.~\ref{sec:sac} and \ref{sec:BIC}.
Next, we send $\mathbf{F}_t$ and $\mathbf{F}_{t+\delta}$ into the shared prediction head to obtain the shadow detection results $\hat{\mathbf{Y}}_t$ and $\hat{\mathbf{Y}}_{t+\delta}$, which are supervised by the ground-truth masks ${\mathbf{Y}}_t$ and $\mathbf{Y}_{t+\delta}$.
Note that the SC-Cor module is flexible and is only used in the training stage without any extra parameters introduced in the test stage, as shown in Fig.~\ref{fig:architecure}~(b). Therefore, it can serve as a plug-and-play component and can be used in many single-image or video shadow detection methods. 

%
\subsection{Shadow-Consistent Correspondence}
\label{sec:sac}

\begin{figure}[t]
	\centering
	\includegraphics[width=0.8\columnwidth,height=0.3\textheight]{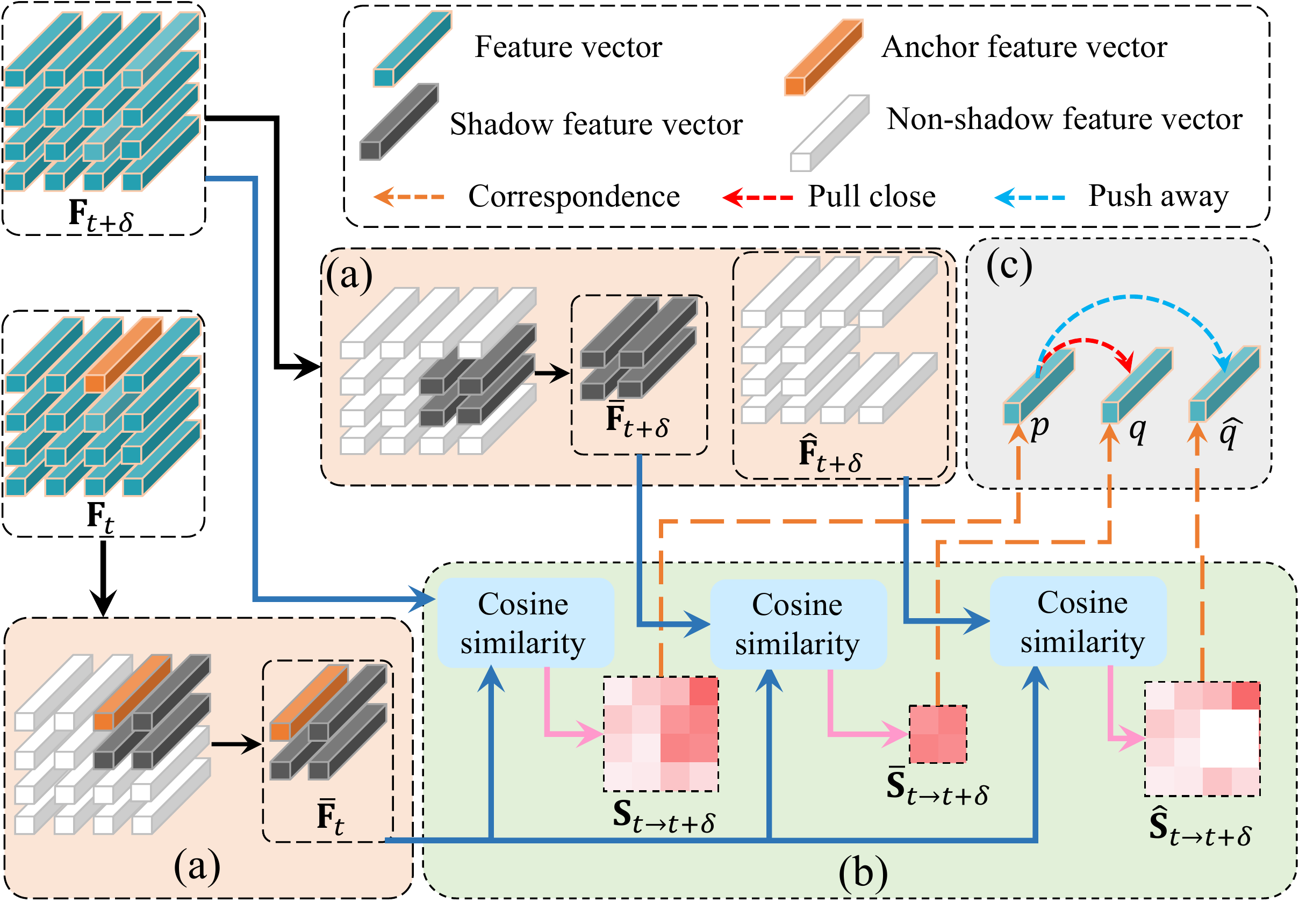}
	\caption{
		Procedure of shadow-consistent correspondence from frame $t$ to $t+\delta$.
		The proposed method consists of three modules: (a) the shadow guidance module (Sec.~\ref{sg}~(a)), (b) the cross-frame correspondence module (Sec.~\ref{cross-f}~(b)), and (c) the consistency regularization (Sec.~\ref{cr}~(c)).
		%
	}
	\label{fig:sac}
\end{figure}

%
To explore the temporal consistency for VSD, we aim to learn shadow correspondence to capture the pixel-wise relations between shadows across frames in the video, which acts as a regularizer to optimize the framework.
%
%
As discussed in Sec.~\ref{sec:intro}, instead of the dense \emph{pixel-to-pixel} labels~\cite{sarlin2020superglue,bian2017gms}, in this paper, we only obtain the \emph{pixel-to-set} labels.
%
To learn dense shadow correspondence, we introduce a novel shadow consistent correspondence method. 
%
%
%
The proposed shadow-consistent correspondence contains three modules: (a) a shadow guidance module, (b) a cross-frame correspondence module, and (c) a consistency regularization, as shown in Fig.~\ref{fig:sac}.

\vspace{1.5mm}
\noindent{\bf (a) Shadow guidance module.}
\label{sg}
%
The shadow guidance module aims to obtain a feature map that only contains feature vector on the shadow regions.
%
Let $\mathbf{F}_t \in \mathbb{R}^{H \times W \times D}$ be the feature map of the frame $V_t$, where $D$, $H$ and $W$ denote the dimension, height, and width of the feature map, respectively.
%
%
Here, we define the ground-truth shadow mask as $\mathbf{Y}_t \in \mathbb{R}^{H \times W \times D}$ and $\mathbf{Y}_t  = \{ 0, 1\}$, where $\mathbf{Y}_t(h,w)=0$ indicates that the position $(h,w)$ in $\mathbf{Y}_t$ is in the non-shadow regions and $\mathbf{Y}_t(h,w)=1$ represents the position $(h,w)$ is in the shadow regions.
%
Then, we can obtain the set of shadow indexes $\mathcal{O}$:
\begin{equation}
	\mathcal{O} \  = \ \{ (h,w) \ | \ \mathbf{Y}_t(h,w)=1 \} \ .
	\label{e:shadow_set}
\end{equation}
\begin{equation}
	\overline{\mathbf{F}}_t  \ = \ \{ \mathbf{F}_t(h,w) \ | \ (h,w) \in \mathcal{O} \}, 
	\label{e:shadow_map}
\end{equation}
where $\overline{\mathbf{F}}_t \in \mathbb{R}^{N_t \times D}$ and $N_t$ indicates the number of shadow feature vectors on $\mathbf{F}_t$,~\emph{i.e.}, $| {\mathcal{O}} | = N_{t}$. We define the operation of the shadow guidance module as $\overline{\mathbf{F}}_t = \text{SG}( \mathbf{F}_t)$.

\vspace{1.5mm}
\noindent{\bf (b) Cross-frame correspondence module.}
\label{cross-f}
%
For each shadow feature vector of a frame, cross-frame correspondence module aims to find its correspondence feature vector,~\emph{i.e.}, most relevant one, from another frame.
%
%
Formally, we define the features of two frames from a video as $\mathbf{F}_t$ and $\mathbf{F}_{t+\delta}$, where $\delta > 0$ is the time interval.
In the following, we will illustrate how to find the correspondence from $\mathbf{F}_t$ to $\mathbf{F}_{t+\delta}$, as well as in the other way around,~\emph{i.e.}, from $\mathbf{F}_{t+\delta}$ to $\mathbf{F}_t$.


%
To find the most correlated feature vector from $\mathbf{F}_{t+\delta}$, we first obtain the shadow feature map of $\mathbf{F}_t$ by $\overline{\mathbf{F}}_t = \text{SG}( \mathbf{F}_t)$.
Then, we measure the similarity between $\overline{\mathbf{F}}_t$ and $\mathbf{F}_{t+\delta}$ by the cosine similarity:
\begin{equation}
	\mathbf{S}_{t \rightarrow t+\delta}  \ = \  \frac{\overline{\mathbf{F}}_t \cdot \mathbf{F}_{t+\delta}}{\| \overline{\mathbf{F}}_t \| \| \mathbf{F}_{t+\delta} \|} \ ,
	\label{e:sim}
\end{equation}
where $\mathbf{S}_{t \rightarrow t+\delta} \in \mathbb{R}^{N_t \times L}$ and $L=H \times W$.
%
Here, we take the $n$-th ($n \in N_t$) feature vector on $\overline{\mathbf{F}}_t$ as the anchor vector, and compute its most relevant feature vector on ${\mathbf{F}}_{t+\delta}$ based on the similarity map $\mathbf{S}_{t \rightarrow t+\delta}$:
%
\begin{equation}
	p  \ = \ \max_{m}\mathbf{S}_{t \rightarrow t+\delta}(n,m), m \in [1, L] \ ,
	\label{e:corres_p}
\end{equation}
where $p$ is the index of the corresponding location on ${\mathbf{F}}_{t+\delta}$. Note that we perform the same operation to find the corresponding $p$ for each feature vector on $\overline{\mathbf{F}}_t$. 

\noindent{\bf (c) Consistency regularization.}
\label{cr}
%
%
Here, we perform the consistency regularization to enforce the found correspondence inside the ground-truth sets.
Specifically, we pull the anchor feature vector close to the shadow ground-truth on the second frame and push it away from the non-shadow regions on the second frame.
%
More specifically, besides measuring the similarity between $\overline{\mathbf{F}}_t$ and $\mathbf{F}_{t+\delta}$, we additionally compute the similarity between $\overline{\mathbf{F}}_t$ and $\overline{\mathbf{F}}_{t+\delta}$ through Eq.~\ref{e:sim}, which can be defined as $ \overline{\mathbf{S}}_{t \rightarrow t+\delta} \in \mathbb{R}^{N_t \times N_{t+\delta}}$, where $N_{t+\delta}$ is the number of shadow feature vectors on $\mathbf{F}_{t+\delta}$.
Then, for the anchor $n$, we find its correspondence on $\overline{\mathbf{F}}_{t+\delta}$ based on $\overline{\mathbf{S}}_{t \rightarrow t+\delta}$ in the same way as Eq.~\ref{e:corres_p}, and we denote the found corresponding location as $q$.
%
%
Note that $p$ is the corresponding location found from the whole feature map while $q$ is the corresponding location only found from the shadow set (indicated by the ground-truth mask).
To pull $p$ and $q$ together, we minimize the discrepancy in two feature similarities by Eq.~\ref{e:shadow_loss}.
\begin{equation}
	\mathcal{L}_\text{shadow}^{t \rightarrow t+\delta } \ = \ \frac{1}{N_t} \sum_{n=1}^{N_t}  \left( \mathbf{S}_{t \rightarrow t+\delta}(n,p) - \overline{\mathbf{S}}_{t \rightarrow t+\delta}(n,q) \right)^2 \ .
	\label{e:shadow_loss}
\end{equation}
%
%
%
To push the anchor $p$ away from the non-shadow regions, we first compute the set of non-shadow indexes $\hat{\mathcal{O}}$:
\begin{equation}
	\hat{\mathcal{O}}  \ = \ \{ (h,w) \ | \ \mathbf{Y}_{t+\delta}(h,w)=0 \} \ .
\end{equation}
%
%
%
Based on $\hat{\mathcal{O}}$, we obtain the non-shadow feature $\hat{\mathbf{F}}_{t+\delta}$.
Then, we compute the similarity between $\overline{\mathbf{F}}_t$ and $\hat{\mathbf{F}}_{t+\delta}$ in the same way as Eq.~\ref{e:sim} to obtain $\hat{\mathbf{S}}_{t \rightarrow t+\delta} \in \mathbb{R}^{N_t \times M_{t+\delta}} $, where $M_{t+\delta}$ is the number of non-shadow features on $\mathbf{F}_{t+\delta}$,~\emph{i.e.}, $| \hat{\mathcal{O}} | = M_{t+\delta}$.
For the anchor $n$, we find its correspondence on $\hat{\mathbf{F}}_{t+\delta}$ based on $\hat{\mathbf{S}}_{t \rightarrow t+\delta}$ in the same way as Eq.~\ref{e:corres_p}, which is denoted as $\hat{q}$.
To push away $p$ and $\hat{q}$, we maximize the margin in two feature similarities by the following loss function:
\begin{equation}
	\begin{aligned}
		&\mathcal{L}_\text{n-shadow}^{t \rightarrow t+\delta }   = \\   & \frac{1}{N_t} \sum_{n=1}^{N_t}   \max \left( 0, 
		\beta - | \mathbf{S}_{t \rightarrow t+\delta}(n,p) - \hat{\mathbf{S}}_{t \rightarrow t+\delta}(n,\hat{q}) | \right)  \ ,
	\end{aligned}
	\label{e:non_shadow_loss}
\end{equation}
where $\beta$ controls the margin between $\mathbf{S}_{t \rightarrow t+\delta}(n,p)$ and $\hat{\mathbf{S}}_{t \rightarrow t+\delta}(n,\hat{q})$.
In the same way, we can obtain the consistency regularization in the other way around, \emph{i.e.}, from $\mathbf{F}_{t+\delta}$ to $\mathbf{F}_{t}$, and we define the loss functions as $\mathcal{L}_\text{shadow}^{t+\delta \rightarrow  t}$ and $\mathcal{L}_\text{n-shadow}^{t+\delta  \rightarrow t}$.
Finally, shadow-consistent correspondence learning can be formulated as follows:
\begin{equation}
	\mathcal{L}_\text{sc} \ = \ \mathcal{L}_\text{shadow}^{t \rightarrow t+\delta } + \mathcal{L}_\text{n-shadow}^{t \rightarrow t+\delta } + \mathcal{L}_\text{shadow}^{t+\delta \rightarrow  t} + \mathcal{L}_\text{n-shadow}^{t+\delta  \rightarrow t} \ .
	\label{e:sac}
\end{equation}

\subsection{Brightness-invariant Correspondence}
\label{sec:BIC}
Due to the changes of brightness in a video, 
the shadow and non-shadow regions in different frames may present similar appearance.
%
To learn brightness-invariant correspondence, we randomly shift the brightness of one frame and learn the brightness-invariant shadow consistency between the shifted frame and another frame in the same video~\cite{zhu2021mitigate}.
Formally, for two frames in a video,~\emph{i.e.}, $V_t$ and $V_{t+\delta}$, we randomly shift the intensity of $V_{t+\delta}$ to produce the shifted frame $V^{\prime}_{t+\delta}=V_{t+\delta}+\gamma$, where $\gamma \in [-\Delta, \Delta]$ is a randomly generated shift parameter and $\Delta$ is a hyper-parameter to control the shift range. 
Next, we use the shadow-consistent learning to learn the cross-frame correspondence between $V_t$ and $V^{\prime}_{t+\delta}$, as introduced in Sec.~\ref{sec:sac}.
\subsection{Overall Objective}
The overall objective of our framework is defined as:
\begin{equation}
	\mathcal{L} \ = \ \mathcal{L}_\text{seg} + \lambda \mathcal{L}_\text{sc} \ ,
	\label{e:loss}
\end{equation}
where $\lambda$ is a hyper-parameter to control the trade-off between these two losses and $\mathcal{L}_\text{seg}$ is the segmentation loss to supervise the pixel-wise prediction.
%
The segmentation loss is different for different shadow detection works.
For example,~\cite{zheng2019distraction,zhu2021mitigate,chen2021triple}
adopt the binary cross entropy (BCE) loss as the segmentation loss:
\begin{equation}
	\mathcal{L}_\text{seg} \ = \ -\frac{1}{T} \sum_{t=1}^{T} \mathbf{Y}_t \cdot \log \left( \hat{\mathbf{Y}}_t \right)+\left(1-\mathbf{Y}_t\right) \cdot \log \left(1-\hat{\mathbf{Y}}_t\right),
\end{equation}
where $\mathbf{Y}_t$ and $\hat{\mathbf{Y}}_t$ are ground-truth and predicted shadow masks, respectively.
%
%
TVSD-Net~\cite{chen2021triple} uses BCE combined with a lovász-hinge loss~\cite{berman2018lovasz}.
%
In the experiments, to highlight the effectiveness of the proposed shadow-consistent correspondence, the segmentation loss keeps consistent with existing papers~\cite{zheng2019distraction,chen2021triple}. 

\section{Experimental Results}
\subsection{Evaluation Metrics and Datasets.}
\label{sec:metrics}
\noindent{\bf Temporal stability.}
Compared with the previous works that only evaluates the performance on each single image (frame-level), in this paper, we introduce a new evaluation metric to evaluate the temporal stability across the video frames, motivated by~\cite{lai2018learning,varghese2020unsupervised}.
In detail, different from \cite{lai2018learning,varghese2020unsupervised} that compute the optical flow between RGB frames, we calculate the optical flow between the ground-truth labels of two adjacent frames,~\emph{i.e.}, $\mathbf{Y}_t$ and $\mathbf{Y}_{t+1}$ through ARFlow~\cite{liu2020learning}, since the motions of shadows are hard to be captured on the RGB frames.
For instance, the optical flows generated by RGB are focus on objects, which can not capture shadows
since the motions of shadows are hard to be captured on the RGB frames; see {\color{black}{supplementary materials}} for more details.
Then, assume $I_{t \rightarrow t+1}$ as the optical flow between $\mathbf{Y}_{t}$ and $\mathbf{Y}_{t+1}$, and we define the reconstructed result that warps $\hat{\mathbf{Y}}_{t+1}$ by the optical flow $I_{t \rightarrow t+1}$ as $\mathbb{Y}_{t}$.
%
Next, we measure the temporal stability of VSD for a video based on the flow warping IoU between the adjacent frames as:
\begin{equation}
	\text{TS} \ = \ \frac{1}{T-1}\sum_{t=1}^{T-1} IoU(\hat{\mathbf{Y}}_{t}, \mathbb{Y}_{t}) \ .
\end{equation}
%

\noindent{\bf Frame-level accuracy.}
Except using the proposed evaluation metric to measure temporal stability, we follow the previous works~\cite{chen2021triple,zheng2019distraction,zhu2021mitigate} and adopt four common evaluation metrics that have been widely used in image/video shadow detection to evaluate the detection accuracy in frame-level.
Specifically, they are Mean Absolute Error (MAE)~\cite{chen2021triple}, F-measure ($F_\beta$)~\cite{chen2021triple,hu2021revisiting}, Intersection over Union
(IoU)~\cite{chen2021triple}, and Balance Error Rate (BER)~\cite{hu2018direction,zhu2021mitigate}.

%
\noindent{\bf Evaluation dataset.}
We conduct our experiments on the ViSha dataset~\cite{chen2021triple} to evaluate the performance.
ViSha consists of $11,685$ image frames and $390s$ duration, which is adjusted to $30$ fps for all video sequences.
This dataset is split into $50$ videos for training and $70$ videos for testing.
\begin{table}[t]
	\caption{{\bf Comparison with the state-of-the-art methods.} ``$\downarrow$'' indicates the lower the scores, the better the results, while ``$\uparrow$'' indicates the higher the scores, the better the results. ``AVG'' is the average score of IoU and TS, which presents the frame-level and temporal-level IoUs. 
		``ISD'' and ``VSD'' stand for the single-image shadow detection and video shadow detection, respectively. ``SOD'' stands for salient object detection. ``VOS'' stands for video object segmentation. ``S-BER'' and ``N-BER'' stand for BER of shadow regions and non-shadow regions, respectively.}
	\centering
	\resizebox{1\columnwidth}{!}{
		\begin{tabular}{l | l | c c c c c c | c | c}
			& & \multicolumn{6}{c|} { \footnotesize \it Frame-level } & { \footnotesize \it Temporal-level}  & \\
			Task &  Method   & MAE  $\downarrow$ & F$_{\beta}$ $\uparrow$ & BER $\downarrow$ & S-BER $\downarrow$ & N-BER $\downarrow$ & IoU [$\%$] $\uparrow$ & TS [$\%$] $\uparrow$ & AVG $\uparrow$ \\
			\Xhline{2.0pt}
			\multirow{2}{*}{Scene Parsing} & FPN~\cite{lin2017fpn}  & 0.044 & 0.707 &  19.49 &	36.59 &	2.40 & 51.28  & 74.27 & 62.78\\
			& PSPNet~\cite{zhao2017pspnet} & 0.052 & 0.642 & 19.75 &	36.44   &3.07 & 47.65 & 76.63 & 62.14 \\
			\Xhline{1.0pt}
			\multirow{1}{*}{SOD} & DSS~\cite{hou2019dss} &  0.045 & 0.697 & 19.78 & 36.96  & 2.59& 50.28 & 75.02 & 62.65 \\
			\Xhline{1.0pt}
			\multirow{5}{*}{VOS} &  PDBM~\cite{song2018pdbm} & 0.066 &0.623  &19.74 &	34.32 &	5.16 &	46.65 & 80.00 & 63.33 \\
			&FEELVOS~\cite{voigtlaender2019feelvos} & 0.043 &	0.710 &	19.76 &	37.27 &	2.26 & 51.20 & 74.89 & 63.05 \\
			&STM~\cite{oh2019stm} & 0.064 & 0.639  &	23.77 &	43.88 &	3.65& 44.69 & 75.30 & 60.00 \\
			\Xhline{1.0pt}
			\multirow{5}{*}{ISD}  &BDRAR~\cite{zhu2018bidirectional}  & 0.050 & 0.695& 21.30 & 40.28 & 2.31 & 48.39  & 72.63 & 60.51\\
			&MTMT~\cite{chen2020multi} &   0.043 &	0.729 &	20.29 &	38.71 &	1.86 &	51.69 & 74.44 & 63.07\\
			&FSD~\cite{hu2021revisiting} & 0.057 &	0.671  &	20.57 &	38.06 &	3.06 &	48.56 & 74.88 &  61.72\\
			\cline{2-10}
			&DSD~\cite{zheng2019distraction} & 0.044 & 0.702 &	19.89 &	37.89 &	1.88 & 51.89 & 74.68&63.29 \\
			&DSD + ours  & 0.039 &0.730 &  15.15 & 27.78 & 2.52 & 58.40 & 78.03 & 68.22\\
			& - & \color{NavyBlue}\textbf{+0.05} & \color{NavyBlue}\textbf{+0.028} & \color{NavyBlue}\textbf{+4.65}&
			\color{NavyBlue}\textbf{+10.11} & \color{Peach}\textbf{-0.64} &  \color{NavyBlue}\textbf{+6.51}&\color{NavyBlue}\textbf{+3.35} & \color{NavyBlue}\textbf{+4.93}\\
			\Xhline{1.0pt}
			\multirow{3}{*}{VSD} & Hu~\emph{et al.}~\cite{hu2021temporal} & 0.078 & 0.683 & 17.03 & 30.13 &  3.93& 51.03 & 83.67&67.35\\
			\cline{2-10}
			&TVSD~\cite{chen2021triple} & 0.033 &	0.757  &17.70 &	33.97 &	1.45&56.57 & 78.25 &67.41 \\
			&TVSD + ours & 0.042 & 0.762 & 13.61 & 24.31 & 2.91 & 61.50 & 81.44 & 71.47\\
			& - & \color{Peach}\textbf{-0.09} & \color{NavyBlue}\textbf{+0.005} & \color{NavyBlue}\textbf{+4.09}&
			\color{NavyBlue}\textbf{+9.66} & \color{Peach}\textbf{-0.46} &  \color{NavyBlue}\textbf{+4.93}&\color{NavyBlue}\textbf{+3.19} & \color{NavyBlue}\textbf{+4.06}\\
	\end{tabular}}
	\label{tab:sota}
\end{table}
\subsection{Implementation Details}
Since our framework is a plug-and-play module that can be used in any shadow detectors, we insert our framework into two state-of-the-art methods on single-image shadow detection and video shadow detection, \emph{i.e.}, DSD~\cite{zheng2019distraction} and TVSD~\cite{chen2021triple}, for evaluation.
During the training process of our shadow-consistent correspondence module, $\lambda$ in Eq.~\ref{e:loss} and $\beta$ in Eq.~\ref{e:non_shadow_loss} are set to $10$ and $0.5$, respectively; please refer to Sec.~\ref{sec:ablation} for the analysis of $\lambda$.
%
$\Delta$ is used to control the range of the brightness shift and it is set to $0.3$ following \cite{zhu2021mitigate}.
Note, shifting brightness of frames will change the distribution of the images, resulting in degrading the detection performance~\cite{zhu2021mitigate}.
Hence, we only shift the brightness of the frames after 2,000 training iterations and freeze the batch normalization~\cite{santurkar2018does}; see {\color{black}{supplementary materials}} for more details.

\begin{figure}[t]
\makeatletter\def\@captype{figure}\makeatother
\begin{minipage}{0.48\linewidth}
\centering
	\includegraphics[width=1.0\columnwidth,height=0.22\textheight]{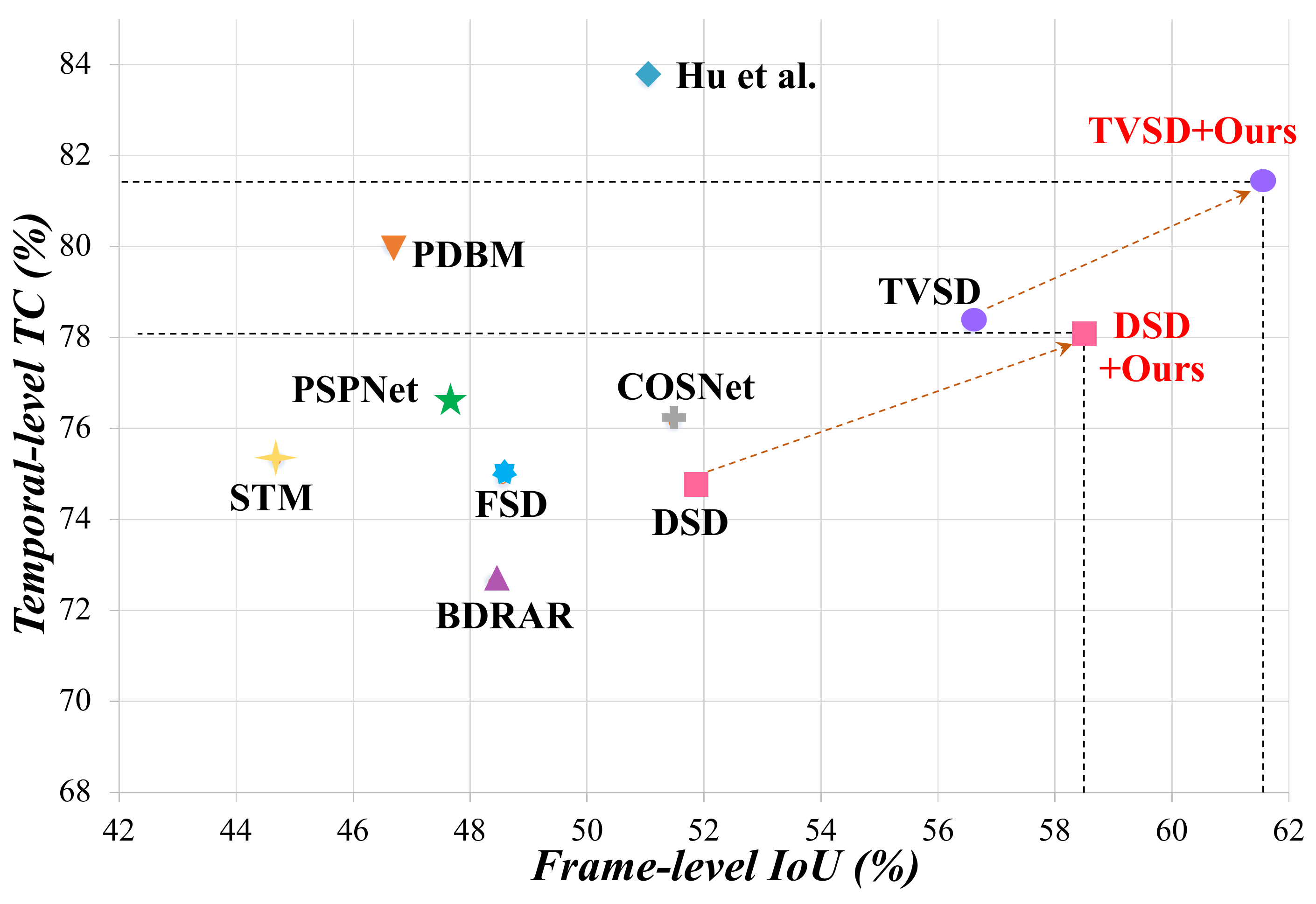}
	\caption{
		{\bf Trade-off between temporal and frame-level accuracy.}
		%
	}
	\label{fig:trade-off}
\end{minipage}
 \hfill
\makeatletter\def\@captype{table}\makeatother
\begin{minipage}{0.50\linewidth}
\centering
\caption{{\bf Comparison of the supervised contrastive learning and ours}.
	Baseline is DSD~\cite{zheng2019distraction}. ``SCon" indicates the supervised contrastive learning. ``Ratio" indicates the ratio of the found correspondence in the shadow regions. ``AVG" refers to the average score of IoU and TS. 
	}
	\resizebox{0.75\columnwidth}{!}{
		\begin{tabular}{   c | c | c }
		   Method & Ratio $\uparrow$ & AVG $\uparrow$ \\
			\Xhline{2.0pt}
		Baseline	&  36.71 & 63.29 \\
		Baseline + \cite{xu2021rethinking} & 82.33 & 66.17 \\
		Baseline + SCon & 85.30 & 65.13\\
		Baseline + Ours & 85.12 & 68.22\\
	\end{tabular}}
	\label{tab:scl&ours}
\end{minipage}
\end{figure}

\begin{figure}[t]
	\centering
	\includegraphics[width=1.0\columnwidth,height=0.18\textheight]{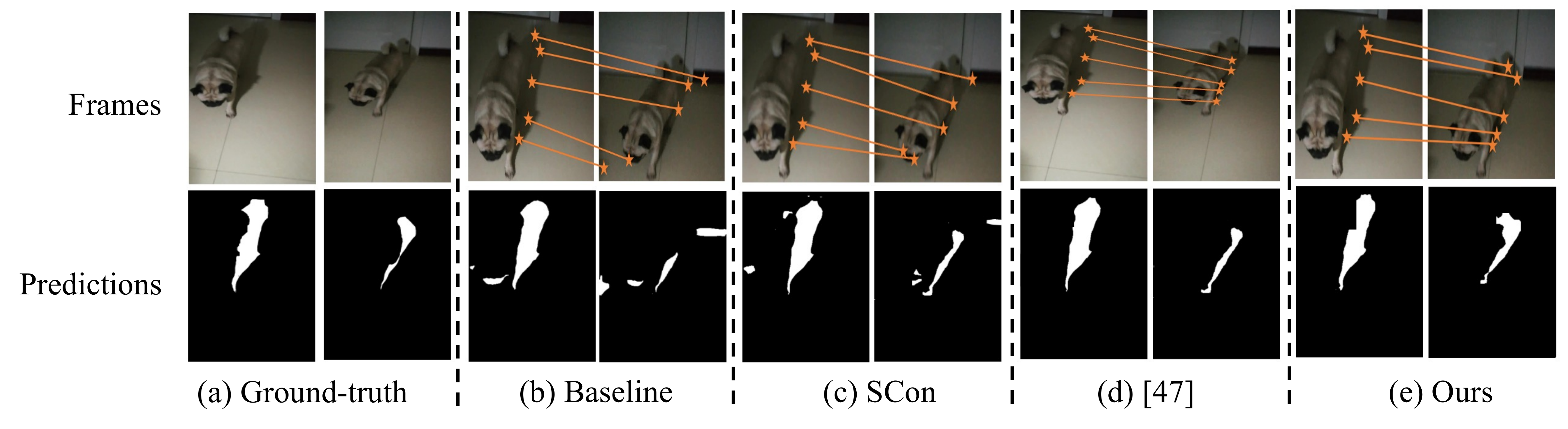}
	\caption{
	    Visualization of the correspondence found by different models.
		Note that Baseline is DSD~\cite{zheng2019distraction}. We sample five pixels in one frame and find their correspondence in the other one.
	}
	\label{fig:corres}
\end{figure}
\begin{figure} [tp]
	\centering
	\includegraphics[width=1.0\columnwidth,height=0.28\textheight]{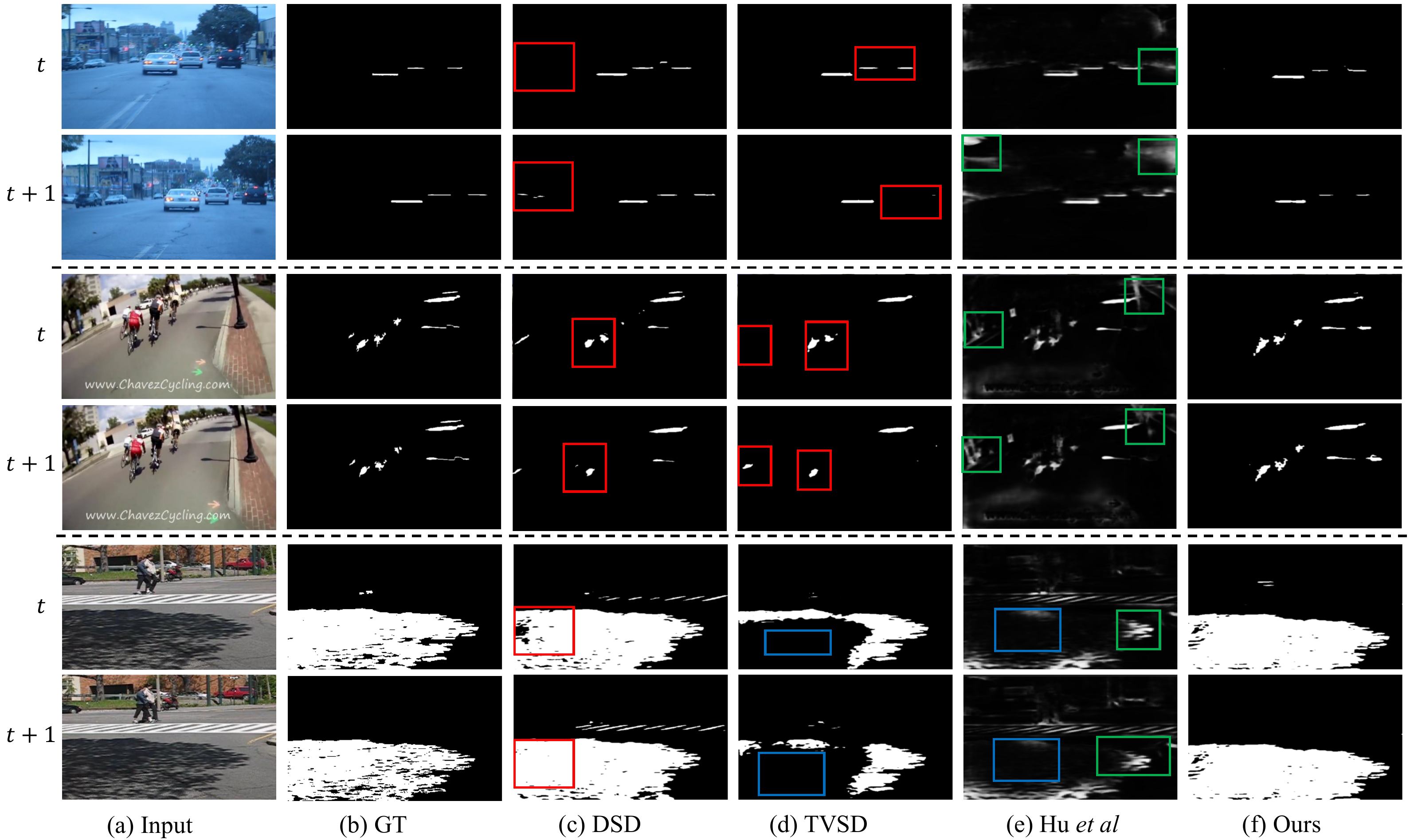}
	\caption{
		Visual comparison of video shadow detection results produced by different methods. (a) is the input images and (b) is the ground-truth (GT) images. (c)-(f) are the results predicted by DSD~\cite{zheng2019distraction}, TVSD~\cite{chen2021triple}, Hu~\emph{et al.}~\cite{hu2021temporal}, and our method, respectively. Our method takes the DSD as the basic network. Note that \textcolor{red}{red boxes} indicate the inconsistent predictions across video frames, \textcolor{blue}{blue boxes} indicate the inaccurate static predictions, and \textcolor{green}{green boxes} show the blurry predictions.
	}
	\label{fig:qualitive}
\end{figure}
\subsection{Comparison with the State-of-the-art Methods}
We conduct the experiments on ViSha~\cite{chen2021triple} to compare with the state-of-the-art methods designed for scene parsing, salient object detection, video object segmentation, single-image shadow detection, and video shadow detection; please see the compared methods in Table~\ref{tab:sota}.
We obtain the results of these methods by retraining them on the ViSha dataset for video shadow detection with the recommendation training parameters or by downloading their results directly from Internet.

Table~\ref{tab:sota} provides the comparison results, which clearly shows that our proposed method can largely improve the performance of both single-image and video shadow detection approaches,~\emph{i.e.}, DSD~\cite{zheng2019distraction} and TVSD~\cite{chen2021triple}, in terms of both frame-level and temporal-level accuracy.
%
%
%
DSD is designed for single-image shadow detection and our approach improves the performance a lot by further considering the dense correspondence among different video frames.
%
Although TVSD is designed for video shadow detection and has explored the temporal consistency in videos, our method further explores the shadow-region correspondence and learns the brightness-invariant features.
%
%
Besides, our method achieves the best trade-off on both temporal-level and frame-level accuracy, as shown in Fig.~\ref{fig:trade-off}.

Fig.~\ref{fig:qualitive} illustrates the visual comparison of the shadow masks produced by DSD~\cite{zheng2019distraction}, TVSD~\cite{chen2021triple}, Hu~\emph{et al.}~\cite{hu2021temporal}, and ours.
From the results, we can see that our method provides more accurate and consistent shadow detection results across different videos frames than others.
More examples and failure cases please refer to {\color{black}{supplementary materials}}.
%

\noindent{\bf Comparison of the supervised contrastive learning, unsupervised correspondence learning and our SC-Cor.}
Specifically, for the sampled pixel of one frame, we find the most correlated pixel in the other frame and denote the found pixel as its correspondence.
%
%
It is clear that the correspondences found by the baseline would be in dark non-shadow regions.
Those found by the supervised contrastive learning would only focus on the shadow regions with similar textures.
For unsupervised correspondence learning, we select \cite{xu2021rethinking}, a recently SOTA, for comparison.
It is clear that ours outperforms [47].
In order to further evaluate the effect of our method, we report the ratio of the found correspondence in ground truth masks and the average performance in Table~\ref{tab:scl&ours}.
The results show that our method can largely improve the accuracy of found correspondence,~\emph{e.g.}, over $43.41\%$ on DSD~\cite{zheng2019distraction}.
Although the ratio of correspondences found by supervised contrastive learning in ground-truth is high, the generated shadow mask may be incomplete; see Fig.~\ref{fig:corres}~(c).  
\subsection{Ablation Study}
\label{sec:ablation}
We conduct ablation experiments to show how each module in our framework design contributes to video shadow detection.
We regard DSD~\cite{zheng2019distraction} as our baseline module in this section.
All the detection results are reported on the testing set of the ViSha dataset~\cite{chen2021triple}.
\begin{table}[tp]
	\caption{{\bf Ablation on the effectiveness of SC-Cor and BS}. ``SC-Cor'' indicates the shadow-consistent correspondence and ``BS'' indicates the brightness shift operation.}
	\centering
	\resizebox{0.85\columnwidth}{!}{
		\begin{tabular}{ l|  c c | c c c | c | c }
&	& &\multicolumn{3}{c|}{ \footnotesize \it Frame-level}&  { \footnotesize \it Temporal-level}  & \\
		Method&	SC & BS & MAE $\downarrow$  & BER $\downarrow$  & IoU [$\%$] $\uparrow$ & TS [$\%$] $\uparrow$ & AVG $\uparrow$ \\
			\Xhline{2.0pt}
	Baseline &		\XSolidBrush & \XSolidBrush & 0.044 &	19.89 & 51.89 & 74.68 & 63.29 \\
 Ours (SC-Cor)	&		\Checkmark  & \XSolidBrush& 0.040 &15.67 &	56.89	&77.09	&67.00 \\
Ours (BS) &		\XSolidBrush   & \Checkmark  &  0.043 & 16.82 & 53.65& 74.92	& 64.29 \\
Ours (Full)	&		\Checkmark &  \Checkmark & {\bf 0.039} & {\bf 14.89} & {\bf 58.40} & {\bf 78.03} & {\bf 68.22}\\
	\end{tabular}}
	\label{tab:sac_bi}
\end{table}
\begin{table}[t]
 \caption{\textbf{Ablation Study on Shadow-consistent correspondence}.}
    \vspace{-2.5mm}
     \begin{subtable}{0.48\columnwidth}
    \caption{{\bf Bidirectional correspondence}. ``Bi-D'' refers to the bidirectional correspondence, which has been defined by $t \rightarrow t +\delta $ and $ t +\delta \rightarrow t$ in Eq.~\ref{e:sac}. }
	\resizebox{1.0\columnwidth}{!}{
		\begin{tabular}{   c  | c c | c | c }
			&\multicolumn{2}{c|}{ \footnotesize \it Frame-level}&  { \footnotesize \it Temporal-level}  & \\
			Bi-D   & BER $\downarrow$  & IoU [$\%$] $\uparrow$ & TS [$\%$] $\uparrow$ & AVG $\uparrow$ \\
			\Xhline{2.0pt}
			\XSolidBrush & 15.54 & 57.25 & 77.08 & 67.17 \\
			\Checkmark & {\bf 14.89} & {\bf 58.40} & {\bf 78.03} & {\bf 68.22} \\
	\end{tabular}}
	\label{tab:bid}
    \end{subtable}
    \begin{subtable}[t]{0.66\columnwidth}
    \end{subtable}
     \begin{subtable}[t]{0.66\columnwidth}
    \end{subtable}
    \vspace{-2.5mm}
     \begin{subtable}{0.52\columnwidth}
      \caption{{\bf Consistency regularization}.``Shadow'' and ``N-Shadow'' indicate the regularization method described in Eq.~\ref{e:shadow_loss} and Eq.~\ref{e:non_shadow_loss}.}
	\resizebox{1.0\columnwidth}{!}{
		\begin{tabular}{   c  | c c | c | c }
			&\multicolumn{2}{c|}{ \footnotesize \it Frame-level}&  { \footnotesize \it Temporal-level}  & \\
			& BER $\downarrow$  & IoU [$\%$] $\uparrow$ & TS [$\%$] $\uparrow$ & AVG $\uparrow$ \\
			\Xhline{2.0pt}
			Shadow &	15.67& 57.18	&76.89 & 67.04\\
			N-shadow	 & 15.79&	57.09&	76.72 & 66.91 \\
			Full	 &{\bf 14.89} & {\bf 58.40} & {\bf 78.03} & {\bf 68.22}\\
	\end{tabular}
	}
	\label{tab:cr}
    \end{subtable}
    \label{tab:ablation}
\end{table}


\paragraph{\bf Effectiveness of SC-Cor and BS.}
%
Table~\ref{tab:sac_bi} reports the effectiveness of the shadow-consistent correspondence (SC-Cor) and the brightness shift (BS) operation.
Training with SC-Cor, we can see a clear improvement in terms of both frame-level accuracy and temporal accuracy,~\emph{i.e.}, $4.22$ on BER and $2.40\%$ on TS.
It is worth noting that only adopting with BS cannot obtain the clear improvement on the temporal stability,~\emph{i.e.}, $74.92\%$ vs. $74.68\%$, due to the lack of exploring temporal information.
By combining with both SC-Cor and BS, the model achieves the best performance.
%

\paragraph{\bf Bidirectional correspondence and consistency regularization.}
Table~\ref{tab:bid} reports the results of bidirectional correspondence in Eq.~\ref{e:sac} and shows the effectiveness of the designed bidirectional correspondence.
Furthermore, we perform the ablation study on the shadow and non-shadow consistency regularization in Table~\ref{tab:cr}, showing that the the combination of them achieves the best results.
\paragraph{\bf Multiple frames and Frame interval.}
We integrate our SC-Cor with multiple pairs of frames in a video and analyze the effectiveness in Table~\ref{tab:multi_frame}.
We observe that training with more frames brings a slight improvement on both frame-level and temporal-level accuracy.
Considering the training efficiency, we choose two pairs of frames.
Furthermore, we study the frame sampling strategy and report the detection results in Table~\ref{tab:frameinterval}.
It is clear that the longer time interval achieves the higher temporal stability while the short one performs better in frame-level accuracy.
For instances, $\delta=1$ achieves the best BER,~\emph{i.e.}, $14.51$, and the lowest TS,~\emph{i.e.}, $76.94\%$.
On the contrary, $\delta=7$ obtains the best TS performance $78.23\%$.
In this paper, we set $\delta$ as five to balance the temporal-level accuracy and the frame-level accuracy.
\begin{table}[t]
 \caption{\textbf{Ablation Study on different frame setting}.}
      \begin{subtable}[t]{0.5\columnwidth}
      \caption{{\bf Multiple frames}. ``Frame number'' denotes the number of sampled frames.}
	\resizebox{1.0\columnwidth}{!}{
		\begin{tabular}{   c  | c c | c | c }
			\multirow{2}{*}{\makecell[c]{Frame \\ number}}   &\multicolumn{2}{c|}{ \footnotesize \it Frame-level}&  { \footnotesize \it Temporal-level}  & \\
			& BER $\downarrow$  & IoU [$\%$] $\uparrow$ & TS [$\%$] $\uparrow$ & AVG $\uparrow$ \\
			\Xhline{2.0pt}
			2    & 15.91 &  58.06 &  77.96 &  68.01\\
			3 & 15.86 &	58.28 &	78.46	& 68.37 \\
			4 & {\bf 15.82} &	{\bf 58.26}	& {\bf 78.51}	 & {\bf 68.39} \\
	\end{tabular}}
	\label{tab:multi_frame}
    \end{subtable}
     \begin{subtable}[t]{0.66\columnwidth}
    \end{subtable}
     \begin{subtable}[t]{0.66\columnwidth}
    \end{subtable}
     \begin{subtable}[t]{0.66\columnwidth}
    \end{subtable}
     \begin{subtable}[t]{0.47\columnwidth}
      \caption{{\bf Frame interval $\delta$}. 
      }
	\resizebox{1.0\columnwidth}{!}{
		\begin{tabular}{   c  | c c | c | c }
			&\multicolumn{2}{c|}{ \footnotesize \it Frame-level}&  { \footnotesize \it Temporal-level}  & \\
			$\delta$ & BER $\downarrow$  & IoU [$\%$] $\uparrow$ & TS [$\%$] $\uparrow$ & AVG $\uparrow$ \\
			\Xhline{2.0pt}
			1&	{\bf 14.51} & {\bf 58.62}& 76.94 & 67.78\\
			3 & 14.73 & 58.55	& 77.23	 & 67.89 \\
			5 & 14.89 & 58.40 &  78.03 & {\bf 68.22} \\
			7 &15.64&	57.12	& {\bf 78.23}	& 67.68\\
	\end{tabular}}
	\label{tab:frameinterval}
    \end{subtable}
    \label{tab:frame}
\end{table}



\section{Conclusion}

In this paper, we present a novel and plug-and-play shadow-consistent correspondence (SC-Cor) method for video shadow detection (VSD). 
A shadow-consistent correspondence is formulated to enforce the network to learn temporal-consistent shadows. A brightness shifting operation is employed to further regularize the network to be brightness-invariant. 
Considering current metrics only evaluate the frame-level accuracy, we introduce a new temporal stability metric, namely TS, for VSD.
Experimental results on the benchmark dataset prove that our SC-Cor outperforms various shadow detection methods.



%
%
\section*{Acknowledgement}
This work was supported by a research grant from HKUST-BICI Exploratory Fund under HCIC-004 and a research grant from Foshan HKUST Projects under FSUST21-HKUST11E.

\clearpage
%
%
\bibliographystyle{splncs04}
\bibliography{egbib}
\end{document}